\documentclass[journal]{IEEEtran}

\usepackage{color}

\usepackage{multirow}

\ifCLASSINFOpdf
  \usepackage[pdftex]{graphicx}
  \graphicspath{{./}}
  \DeclareGraphicsExtensions{.jpg,.png}%,.pdf}
\else
\fi
\usepackage{amsmath}
\usepackage{amsfonts}
\begin{document}

\newpage
\onecolumn

\bigskip
\bigskip
\bigskip
\bigskip
\bigskip
\bigskip

{\Large
This paper is a preprint (IEEE “accepted” status). IEEE copyright notice.  2018 IEEE. Personal use of this material is permitted. Permission from IEEE must be obtained for all other uses, in any current or future media, including reprinting/republishing this material for advertising or promotional purposes, creating new collective works, for resale or redistribution to servers or lists, or reuse of any copyrighted.}

\bigskip
\bigskip
\bigskip
\bigskip
\bigskip
\bigskip

{\Large
A. Tejero-de-Pablos, Y. Nakashima, T. Sato, N. Yokoya, M. Linna and E. Rahtu, "Summarization of User-Generated Sports Video by Using Deep Action Recognition Features," in IEEE Transactions on Multimedia.

doi: 10.1109/TMM.2018.2794265

keywords: {Cameras; Feature extraction; Games; Hidden Markov models; Semantics; Three-dimensional displays; 3D convolutional neural networks; Sports video summarization; action recognition; deep learning; long short-term memory; user-generated video},

\bigskip

URL: http://ieeexplore.ieee.org/stamp/stamp.jsp?tp=\&arnumber=8259321\&isnumber=4456689}
\thispagestyle{empty}\newpage
\twocolumn

%
% paper title
% Titles are generally capitalized except for words such as a, an, and, as,
% at, but, by, for, in, nor, of, on, or, the, to and up, which are usually
% not capitalized unless they are the first or last word of the title.
% Linebreaks \\ can be used within to get better formatting as desired.
% Do not put math or special symbols in the title.
\title{Summarization of User-Generated Sports Video\newline by Using Deep Action Recognition Features}
%
%
% author names and IEEE memberships
% note positions of commas and nonbreaking spaces ( ~ ) LaTeX will not break
% a structure at a ~ so this keeps an author's name from being broken across
% two lines.
% use \thanks{} to gain access to the first footnote area
% a separate \thanks must be used for each paragraph as LaTeX2e's \thanks
% was not built to handle multiple paragraphs
%

\author{Antonio~Tejero-de-Pablos,~\IEEEmembership{Member,~IEEE,}
        Yuta~Nakashima,~\IEEEmembership{Member,~IEEE,}
        Tomokazu~Sato,~\IEEEmembership{Member,~IEEE,}
        Naokazu~Yokoya,~\IEEEmembership{Life Senior Member,~IEEE,}
        Marko~Linna,~%\IEEEmembership{Member,~IEEE,???}
        and~Esa~Rahtu%,~\IEEEmembership{Member,~IEEE}% <-this % stops a space
\thanks{A. Tejero-de-Pablos, Y. Nakashima, T. Sato and N. Yokoya were with the Graduate School of Information Science, Nara Institute of Science and Technology, Nara, 630-0101 Japan. e-mail: antonio.tejero.ao4@is.naist.jp.}% <-this % stops a space
\thanks{M. Linna is with the Center for Machine Vision Research, University of Oulu, Oulu, 90014 Finland. E. Rahtu is with the Department of Signal Processing, Tampere University of Technology, Tampere, 33101 Finland}% <-this % stops a space
\thanks{Manuscript received July, 2017}}%; revised August 26, 2015.}}

% note the % following the last \IEEEmembership and also \thanks - 
% these prevent an unwanted space from occurring between the last author name
% and the end of the author line. i.e., if you had this:
% 
% \author{....lastname \thanks{...} \thanks{...} }
%                     ^------------^------------^----Do not want these spaces!
%
% a space would be appended to the last name and could cause every name on that
% line to be shifted left slightly. This is one of those "LaTeX things". For
% instance, "\textbf{A} \textbf{B}" will typeset as "A B" not "AB". To get
% "AB" then you have to do: "\textbf{A}\textbf{B}"
% \thanks is no different in this regard, so shield the last } of each \thanks
% that ends a line with a % and do not let a space in before the next \thanks.
% Spaces after \IEEEmembership other than the last one are OK (and needed) as
% you are supposed to have spaces between the names. For what it is worth,
% this is a minor point as most people would not even notice if the said evil
% space somehow managed to creep in.

% The paper headers
\markboth{IEEE TRANSACTIONS ON MULTIMEDIA}%,~Vol.~14, No.~8, August~2015}
{Tejero-de-Pablos \MakeLowercase{\textit{et al.}}: Summarization of User-Generated Sports Video by Using Deep Action Recognition Features}
% The only time the second header will appear is for the odd numbered pages
% after the title page when using the twoside option.
% 
% *** Note that you probably will NOT want to include the author's ***
% *** name in the headers of peer review papers.                   ***
% You can use \ifCLASSOPTIONpeerreview for conditional compilation here if
% you desire.

% If you want to put a publisher's ID mark on the page you can do it like
% this:
%\IEEEpubid{0000--0000/00\$00.00~\copyright~2015 IEEE}
% Remember, if you use this you must call \IEEEpubidadjcol in the second
% column for its text to clear the IEEEpubid mark.

% use for special paper notices
%\IEEEspecialpapernotice{(Invited Paper)}

% make the title area
\maketitle

% As a general rule, do not put math, special symbols or citations
% in the abstract or keywords.
\begin{abstract}
Automatically generating a summary of sports video poses the challenge of detecting interesting moments, or highlights, of a game.
%Summarizing user-generated sports video requires doing this in absence of editing conventions (e.g.~superimposed text and specific camera work) that facilitate the extraction of high-level semantics.
Traditional sports video summarization methods leverage editing conventions of broadcast sports video that facilitate the extraction of high-level semantics. However, user-generated videos are not edited, and thus traditional methods are not suitable to generate a summary.
In order to solve this problem, this work proposes a novel video summarization method that uses players' actions as a cue to determine the highlights of the original video. A deep neural network-based approach is used to extract two types of action-related features and to classify video segments into interesting or uninteresting parts. The proposed method can be applied to any sports in which games consist of a succession of actions. Especially, this work considers the case of Kendo (Japanese fencing) as an example of a sport to evaluate the proposed method. The method is trained using Kendo videos with ground truth labels that indicate the video highlights. The labels are provided by annotators possessing different experience with respect to Kendo to demonstrate how the proposed method adapts to different needs. The performance of the proposed method is compared with several combinations of different features, and the results show that it outperforms previous summarization methods.
\end{abstract}

% Note that keywords are not normally used for peerreview papers.

% For peer review papers, you can put extra information on the cover
% page as needed:
% \ifCLASSOPTIONpeerreview
% \begin{center} \bfseries EDICS Category: 3-BBND \end{center}
% \fi
%
% For peerreview papers, this IEEEtran command inserts a page break and
% creates the second title. It will be ignored for other modes.
\IEEEpeerreviewmaketitle

\section{Introduction}

\IEEEPARstart{T}{he} widespread availability of cameras has led to an enormous and ever-growing collection of unedited and unstructured videos generated by users around the world \cite{jiang2015super}. A popular domain corresponds to sports videos taken at public events and professional/amateur matches. These types of user-generated sports videos (UGSVs) are often lengthy with several uninteresting parts, and thus many of them are stored and are never reviewed. A convenient way to review, transfer, and share the video via channels, such as social network services, includes generating summaries of a UGSV that only shows the interesting parts or highlights. 

%On the other hand, research efforts have been made as well for general user-generated videos. Existing methods range from traditional clustering-based one, which eliminates redundancy \cite{lienhart1999dynamic}, to the most recent deep neural network-based one, which learns features that allow to model highlights of video \cite{yang2015unsupervised}.

%Since manually extracting video highlights is time-consuming, the field of video summarization \cite{truong2007video} studies automatic highlight extraction to compact the content of a video.

Automatic video summarization is a challenging problem that involves extracting semantics from video. Traditional user-generated video summarization methods target general videos in which contents are not limited to a specific domain. This is mainly because of the difficulty in extracting semantics from an unstructured video \cite{zha2007building}. As opposed to extracting semantics, these methods use low-level visual features and attempt to reduce visual redundancy using clustering-based approaches \cite{lienhart1999dynamic}. More recent user-generated video summarization methods use deep neural network-based features to extract higher-level semantics \cite{zhang2016video,otani2016videosum}. 

\begin{figure}[!t]
\centering
\includegraphics[width=\columnwidth]{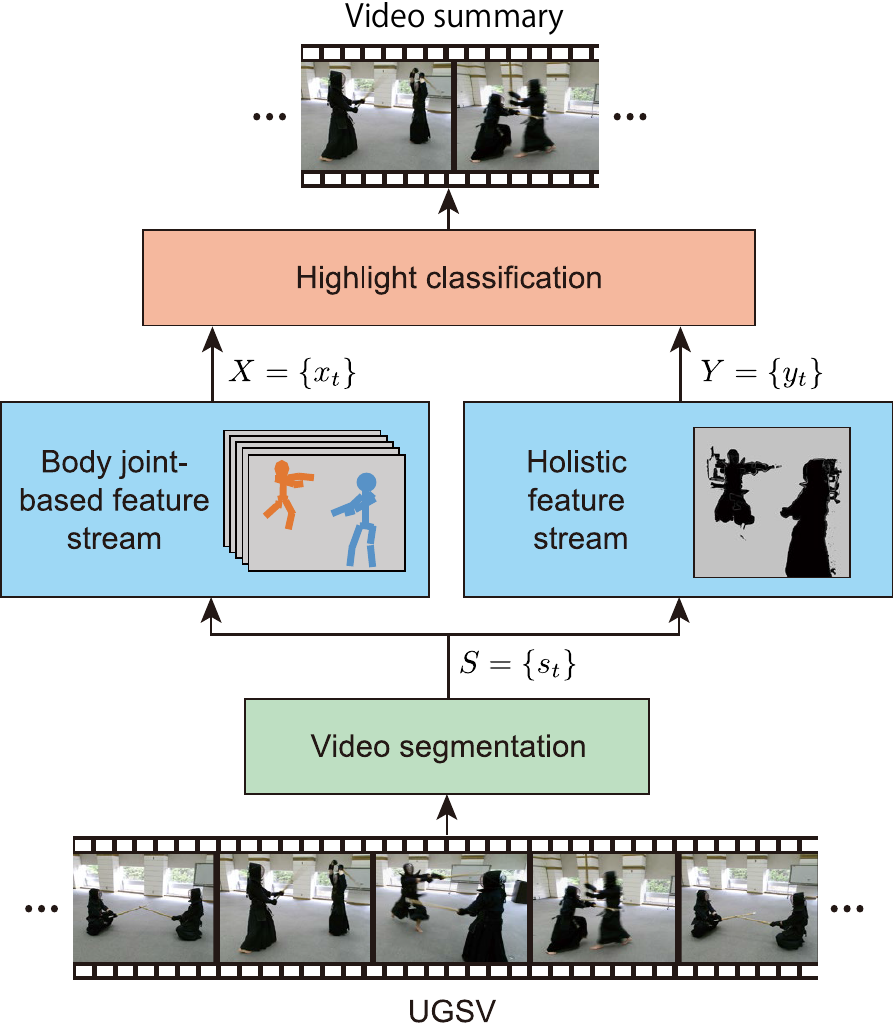}
%%\vspace{-7mm}
\caption{An overview of the proposed method to generate a summary of user-generated sports video (UGSV) based on players' actions. Two types of features that represent players' actions, namely body joint-based and holistic actions, are used to extract highlights from the original video.}
%%\vspace{-4mm}
\label{fig:overview}
\end{figure}

With respect to sports video and especially with respect to professional sports in broadcast TV programs, there exist a number of summarization methods that leverage editing conventions to extract high-level semantics by exploiting a knowledge of the specific sport \cite{li2010action,nitta2009automatic}. For example, broadcast sports video contains slow-motion replays \cite{pan2001detection}, narration and superimposed text \cite{ekin2003automatic}, and specific camera work \cite{xu2005hmm}. This type of video editing constitutes the basis for heuristic rules that aid in the determination of highlights (or certain interesting moments of a game such as a free kick in soccer or a pitch in baseball). Additionally, broadcast video is often edited by following the structure of the sport (i.e., ``downs'' in American football), and this constitutes another cue for summarization \cite{li2001event}.

UGSV lies in in-between general user-generated video and broadcast sports video. Given a specific sport, domain knowledge can be used to generate a UGSV summary. However, UGSV does not typically follow any editing convention or structure, and thus a different type of cues is required to grab the semantics.

This paper describes a novel method for UGSV summarization. Our observation with respect to semantics extraction is that a game in most sports consists of a succession of players' actions, and thus the actions can be one of the most important cues to determine if a certain part of video is interesting or not.
For example, a definitive smash in tennis is more likely to be enjoyed by tennis viewers than a repetitive ball exchange. Also, a feint in boxing might not be interesting by itself, but viewers would surely enjoy it if it is followed by an uppercut that knocks out the opponent.
Based on this observation, the proposed method uses players' actions to model the highlights of a sports game (Fig.~\ref{fig:overview}).

Inspired by recent methods for action recognition in video, the proposed method uses a two-stream architecture that extracts two types of action features for action representation. One type involves players' body joint positions estimated in 2D or 3D (obtainable from depth maps). Body joint-based features provide a precise representation of actions. The other type involves holistic features that can be obtained with deep convolutional neural networks (CNNs) designed to extract spatio-temporal features from video. Holistic features help to capture actions in their context. Subsequently, long short-term memory (LSTM) is used to model the temporal dependencies of the extracted features for highlight classification. In our summaries, a highlight may contain one or more actions performed by the players. Several types of body joint-based and holistic features are comparatively evaluated for UGSV summarization.

We consider the case of Kendo (Japanese fencing) as an example of a sport to evaluate the proposed method. This work is an extension of our previous work in \cite{tejerodepablos2016human}.

The main contributions of this work are as follows:
\begin{itemize}
\item A novel UGSV summarization method that determines highlights in a video by using players' action features and a deep neural network. 
\item A comparison of several action feature extraction methods, i.e., body joint features (RGB image-based 2D joint positions and depth map-based 3D joint positions) and holistic features (C3D \cite{tran2015learning} and CNN-ISA \cite{le2011learning}) to demonstrate their adequacy to model video highlights.
\item A new UGSV dataset with 246 min of Kendo videos in which each second of a video has a label that indicates whether or not it is a part of a highlight. The labels were provided by annotators with and without experience in Kendo. 
\item Objective and subjective evaluations of the proposed method. Users with and without experience in Kendo were surveyed to investigate the adequacy of the proposed method with respect to individual needs.
\end{itemize}

%\textcolor{red}{The remainder of this paper is organized as follows. Section \ref{sec:related} reviews the state-of-the-art methods in video summarization and shows why they cannot be applied to user-generated sports video. Section \ref{sec:harsum} describes our methodology for motion feature extraction and highlights modeling for user-generated sports video. Section \ref{sec:experiments} details the experimental results of evaluating a variety of feature extraction methods. Finally Section \ref{sec:conclusions} draws the main points of this work.}

\section{Related work}
\label{sec:related}

%This section reviews the main state-of-the-art works in the fields of sports video summarization, UGV summarization and human action recognition, with an emphasis on their respective feature extraction methodology.

This section introduces existing video summarization methods in terms of the types of video (i.e., broadcast sports video and user-generated video). This section also reviews existing work in action recognition, which constitutes a key technique for modeling highlights in the proposed method.

\subsection{Action recognition from video}

Body-joint features are widely used for human action recognition, because of their rich representability of human motion and their robustness to variability in human appearance \cite{biswas2011gesture}. However, they miss potential cues contained in the appearance of the scene. Holistic features, which focus more on the global appearance of the scene, have been also hand-crafted for action recognition \cite{sun2009action}; from motion energy-images to silhouette-based images \cite{chen2007human,calderara2008action}. As shown in recent works \cite{tran2015learning,le2011learning,simonyan2014two,feichtenhofer2016spatiotemporal}, convolutional neural networks (CNN) have outperformed traditional methods as they are able to extract holistic action recognition features that are more reliable and generalizable than hand-crafted features.
An example corresponds to three-dimensional convolutional neural networks (3D CNNs) that constitute an extension of CNNs applied to images (2D CNNs). While 2D CNNs perform only spatial operations in a single image, 3D CNNs also perform temporal operations while preserving temporal dependencies among the input video frames \cite{tran2015learning}. Le et al.~\cite{le2011learning} used a 3D CNN with independent subspace analysis (CNN-ISA) and a support vector machine (SVM) to recognize human actions from video. Additionally, Tran et al.~\cite{tran2015learning} designed a CNN called C3D to extract video features that were subsequently fed to an SVM for action recognition.

Another state-of-the-art CNN-based action recognition method employed two types of streams, namely a spatial \textit{appearance} stream and a temporal \textit{motion} stream \cite{simonyan2014two,feichtenhofer2016spatiotemporal}. Videos are decomposed into spatial and temporal components, i.e., into an RGB and optical flow representation of its frames, and fed into two separate 3D CNNs. Each stream separately provides a score for each possible action, and the scores from two streams were later combined to obtain a final decision. This architecture is supported by the two-stream hypothesis of neuroscience in which the human visual system is composed of two different streams in the brain, namely the dorsal stream (spatial awareness and guidance of actions) and the ventral stream (object recognition and form representation) \cite{goodale1992separate}.

In addition to RGB videos, other methods leverage depth maps obtained from commodity depth sensors (e.g.~Microsoft Kinect) to estimate the human 3D pose for action recognition \cite{xia2012view,martinez2014action,cai2016effective}. The third dimension provides robustness to occlusions and variations from the camera viewpoint.

\subsection{Broadcast sports video summarization}

Summarization of sports video focuses on extracting interesting moments (i.e., highlights) of a game. A major approach leverages editing conventions such as those present in broadcast TV programs. Editing conventions are common to almost all videos of a specific sport and allow automatic methods to extract high-level semantics \cite{choi2008spatio,chen2006semantic}. Ekin et al.~\cite{ekin2003automatic} summarized broadcast soccer games by leveraging predefined camera angles in edited video to detect soccer field elements (e.g., goal posts). Similar work used slow-motion replays to determine key events in a game \cite{pan2001detection} and predefined camera motion patterns to find scenes in which players scored in basketball/soccer games \cite{xu2005hmm}.

In addition to editing conventions, the structure of the sport also provides high-level semantics for summarization. Certain sports are structured in ``plays'' that are defined based on the rules of the sport and are often easily recognized in broadcast videos \cite{li2010action,tjondronegoro2004integrating,liang2004semantic}. For example, Li et al.~\cite{li2001event} summarized American football games by leveraging their turn-based structure and recognizing ``down'' scenes from the video. Other methods used metadata in sports videos \cite{nitta2009automatic,divakaran2003video} since it contains high-level descriptions (e.g., ``hits'' may be annotated in the metadata with their timestamps for a baseball game). A downside of these methods is that they cannot be applied to sports video without any editing conventions, structures, and metadata. Furthermore, they are based on heuristics, and thus it is difficult to generalize them to different sports.

Existing work also proposed several methods that are not based on heuristics.
%that does not use high-level semantics
These methods leverage variations between scenes that are found in broadcast video (e.g., the close-up in a goal celebration in soccer). Chen et al.~\cite{chen2008motion} detected intensity variations in color frames to segment relevant events to summarize broadcast videos of soccer, basketball, and tennis. Mendi et al.~\cite{mendi2013sports} detected the extrema in the optical flow of a video to extract the frames with the highest action content and construct a summary for broadcast rugby video. These methods can be more generally applied to broadcast videos, but they lack high-level semantics, and thus the extracted scenes do not always correspond to the highlights of the game.

\subsection{User-generated video summarization}
\label{sec:ugv}

Sports video includes a somewhat universal criterion on the extent to which a ``play'' is interesting (e.g., a \textit{homerun} in a baseball game should be an interesting play for most viewers). In contrast, user-generated video in general do not have a clear and universal criterion to identify interesting moments. Additionally, neither editing conventions nor specific structures that can be used to grab high-level semantics can be leveraged \cite{hua2004optimization}. Hence, many video summarization methods for user-generated video are designed to reduce the redundancy of a lengthy original video as opposed to determining interesting moments. Traditional methods uniformly sample frames \cite{mills1992magnifier} or cluster them based on low-level features, such as color \cite{lienhart1999dynamic}, to extract a brief synopsis of a lengthy video. These methods do not extract highlights of the video, and therefore researchers proposed other types of summarization criteria such as important objects \cite{meng2016keyframes},
%(e.g., the bride in a wedding video \cite{cheng2008semantic})
attention \cite{evangelopoulos2013multimodal}, interestingness \cite{peng2011editing}, and user preferences \cite{garciadelmolino2017active}.%gygli2014creating

Recent methods use deep neural networks to automatically learn a criterion to model highlights. Yang et al.~\cite{yang2015unsupervised} extracted features from ground-truth video summaries to train a model for highlight detection. Otani et al.~\cite{otani2016video} use a set of both original videos and their textual summaries that are generated via majority voting by multiple annotators to train a model to find video highlights. Video titles \cite{song2015tvsum}, descriptions \cite{otani2016videosum}, and other side information \cite{yuan2017video} can also be used to learn a criterion to generate summaries. The aforementioned methods employed networks with CNNs and LSTMs, and this requires a large amount of data for training. The generation of these types of large summarization datasets for training their network is non-viable for most researchers, and thus their models are built on pre-trained networks such as VGG \cite{simonyan2014very} and GoogLeNet \cite{szegedy2015going}.
%\tc{These networks lack any kind of motion modeling, focusing only in the appearance of certain objects, which may be too generic to represent a video of a specific sport. However, in a genre such as sports video, motion plays a very important role, so human motion should be also considered when extracting features to model video highlights. (why is this block necessary?)}

\section{UGSV summarization using action features}
\label{sec:harsum}

UGSV summarization inherits the intricacies of user-generated video summarization. The extraction of high-level semantics is not trivial in the absence of editing conventions. However, given a specific sport, it is possible to leverage domain knowledge to facilitate the extraction of high-level semantics. The idea in the present work for semantics extraction involves utilizing players' actions, as they are the main constituents of a game. Our previous work \cite{tejerodepablos2016human} applied an action recognition technique to sports videos to determine combinations of actions that interest viewers by using a hidden Markov model with Gaussian Mixture emissions. To the best of our knowledge, this work was the first to use a UGSV summarization based on players' actions. 

A major drawback of the previous work \cite{tejerodepablos2016human} involves the usage of the outputs of a classic action recognizer as features to determine the highlights of the UGSV. Moreover, in addition to the UGSV summarization dataset, the method also requires an action dataset of the sport to train the action recognizer.
%The recent trend of deep neural networks has demonstrated the power of feature learning, in which a neural network is trained in an end-to-end manner from its input to the top layers or at least partially from one of its layers to the top.
Another drawback of \cite{tejerodepablos2016human} is that it only uses features from 3D joint positions (that are estimated by, e.g.,~\cite{zhang2012microsoft}). They provide rich information on players' actions but miss other potential cues for summarization contained in the scene. Holistic features can compensate such missing cues by, for example, modeling the context of an action. Also, holistic features are useful when the joint position estimation fails.

Hence, in this work, we hypothesize that features extracted from players' actions allow summarizing UGSV. The method in \cite{tejerodepablos2016human} is extended by employing a two-stream deep neural network \cite{simonyan2014two, feichtenhofer2016spatiotemporal}. Our new method considers two different types of inputs, namely RGB frames of video and body joint positions, and each of them are transformed through two separate neural networks (i.e., streams). These two streams are then fused to form a single action representation to determine the highlights. Our method does not require recognizing the actions explicitly, thus avoiding expensive human action annotation; and the proposed network is trained from the lower layer to the top layers by using a UGSV summarization dataset.

Given the proposed method, it is necessary for the target sports to satisfy the following conditions: (1) a game consists of a series of recognizable actions performed by each player and (2) players are recorded from a close distance for joint position estimation. However, it is expected that the idea of using action recognition-related features for UGSV summarization is still valid for most types of sports.

\subsection{Overview}

In this work, UGSV summarization is formulated as a problem of classifying a video segment in the original video as interesting (and thus included in the summary) or uninteresting. A two-stream neural network is designed for this problem, and it is trained in a supervised manner with ground truth labels provided by multiple annotators.

Figure \ref{fig:overview} shows an overview of the proposed method. The method first divides the original input video into a set $S = \{s_t\}$ of video segments in which RGB frames can be accompanied by their corresponding depth maps. A video segment $s_t$ is then fed into the two-stream network. The body joint-based feature stream considers RGB frames (and depth maps) in $s_t$ to obtain body joint-based features $x_t$, and the holistic feature stream computes holistic features $y_t$ from the RGB frames. The former stream captures the players' motion in detail by explicitly estimating their body joint positions. The latter stream represents entire frames in the video segment, and this is helpful to encode, for example, the relationship between the players. The features $X = \{x_t\}$ and $Y = \{y_t\}$ are then used for highlight classification by considering the temporal dependencies among the video segments. The highlight summaries correspond to a concatenation of the segments that are classified as interesting.

\subsection{Video segmentation}

\begin{figure}[!t]
\centering
\includegraphics[width=\columnwidth]{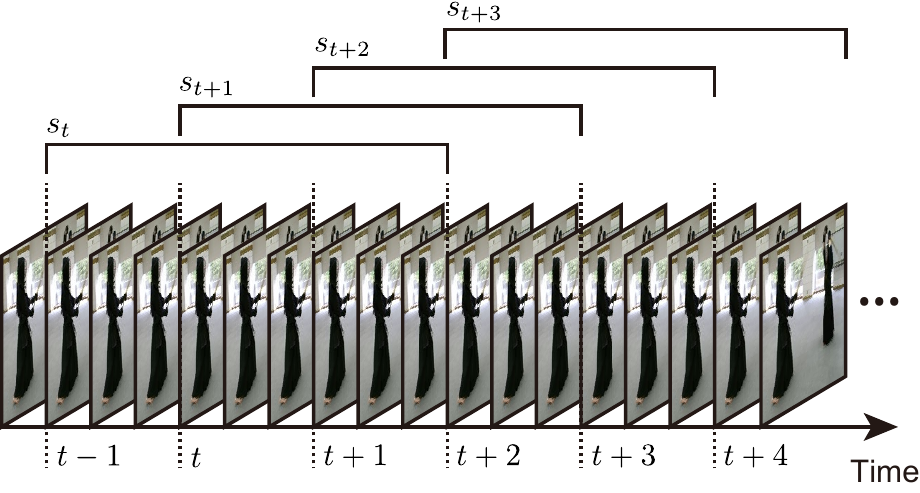}
\caption{In the video segmentation, a video segment $s_t$ contains frames in-between $t-1$ and $t+2$ sec. Each video segment overlaps with adjacent ones for 2 sec.}
\label{fig:segmentation}
\end{figure}

Various methods have been proposed to segment a video (e.g., its content \cite{chen2008motion}). In the proposed method, the original input video of length $T$ seconds (sec) is uniformly segmented into multiple overlapping segments, so that subsequently a second $t$ of video can be represented by extracting action features from a video segment $s_t$, i.e., $S = \{s_t| t=1,\dots,T\}$. Thus, $T$ also corresponds to the number of video segments in $S$, and $s_t$ corresponds to the video segment that contains frames from sec $t-1$ to sec $t+\tau-1$. For a finer labeling of highlights, short video segments are required. We choose a $\tau = 3$, for which adjacent video segments overlap by 2 sec as shown in Fig.~\ref{fig:segmentation}. Each segment $s_t$ may contain a different number of frames, especially when the input video is captured with an RGB-D camera (e.g., Microsoft Kinect) due to automatic exposure control.

\subsection{Body joint-based feature stream}
\label{sec:jointfeat}

\begin{figure}[!t]
\centering
\includegraphics[width=\columnwidth]{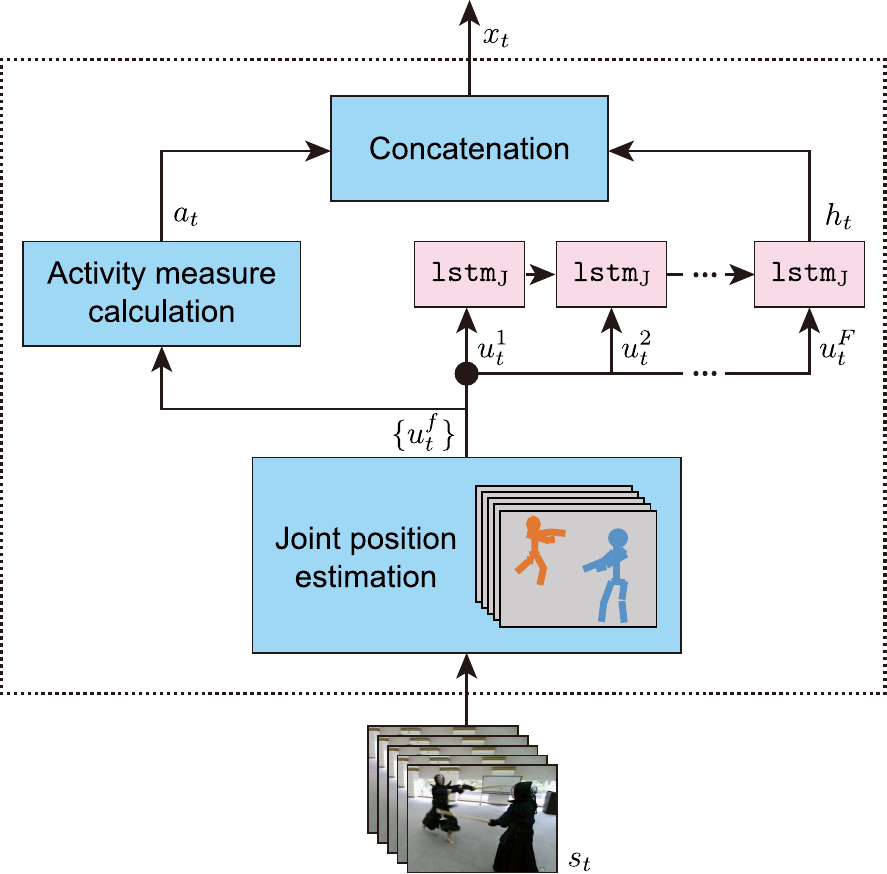}
\caption{In the body joint-based feature stream, an LSTM is fed with the body joint positions estimated from players on each frame $u_t^f$ to model temporal dependencies and extract a feature vector $h_t$. Additionally, these body joint positions are also used to calculate an activity measure for all players $a_t$. The body joint-based feature vector is their concatenation $x_t$.}
\label{fig:jointfeat}
\end{figure}

In this stream (Fig.~\ref{fig:jointfeat}), a sequence of positions of the players' body joints (e.g., head, elbow, etc.) that represent the movement of the players irrespective of their appearance is used to obtain a detailed representation of players' actions. Specifically, two types of joint representations are employed in this work, namely 3D positions from depth maps or 2D positions from RGB frames. 

With respect to the 3D body joint positions, the skeleton tracker (e.g., \cite{zhang2012microsoft}) is used as in the previous work \cite{tejerodepablos2016human}, and it estimates 3D positions from depth maps. The 3D positions are usually represented in the camera coordinate system, and thus they are view-dependent, thereby introducing extra variations. Therefore, the 3D positions from the camera coordinate system are transformed to each player's coordinate system in which the origin corresponds to one of the body joints (e.g., torso). 

In the absence of depth maps (which is likely in current user-generated video), 2D body joint positions can still be estimated from RGB frames. Recent methods in human pose estimation leverage 2D CNNs to learn spatial relationships among human body parts and estimate 2D joint positions \cite{wei2016convolutional}. These types of 2D positions are not as robust relative to view variations as 3D positions. However, they can be extracted from RGB frames alone without using depth maps. The given 2D body joint positions are also transformed to positions relative to the player's coordinate system to ensure that they are translation invariant.

The use of an activity measure works positively while extracting highlights \cite{tejerodepablos2016human}. In order to calculate the activity measure of a certain player $q$ in the video segment $s$ (we omit subscript $t$ in this subsection for notation simplicity), the volume (or plane for the 2D case) around the player is divided into a certain number of regions, and the ratio $r_v$ of the number of frames in the video segment in which the joint $j$ falls into region $v$ is calculated. The activity measure $a_q$ is defined as the entropy obtained based on $r_v$. With respect to each joint $j$ in player $q$'s body, we compute the entropy as follows:
\begin{equation}
e_j = -\sum_v r_v \log(r_v).
\end{equation}
Then, we calculate the activity measure for player $q$ as follows:
\begin{equation}
a_q = \sum_{j=1}^J e_j.
\end{equation}
The activity measure for all players in a segment is calculated. More details on the activity measure can be found in \cite{tejerodepablos2016human}.

Let $u^f_{qj} $ in $\mathbb{R}^3 $ or $\mathbb{R}^2$ (a row vector) denote the 3D or 2D relative position of joint $j$ of player $q$ in frame $f$ of video segment $s$. Subsequently, given the number of players $Q$ and of estimated body joints $J$, the concatenation of the body joints of all players in frame $f$ is defined as follows:
\begin{equation}
u^f  = (u^f_{11} \cdots u^f_{qj} \cdots u^f_{QJ}).
\end{equation}
As shown in Fig.~\ref{fig:jointfeat}, vectors $u^1$ to $u^F$ are passed through an LSTM to model the temporal dependencies of the joint positions of players' bodies in $s$. After feeding the last vector $u^F$, the hidden state vector $h$ of the LSTM is considered as a representation of $\{u^f\}$. The state of the LSTM is reset to all zeros prior to feeding the next video segment. It is assumed that the number of players $Q$ does not change. However, some players can be out of the field-of-view of the camera. In this case, the corresponding elements in $u$ and $a$ are substituted with zeros.

The proposed method represents a video segment $s$ by concatenating the LSTM output and the activity measure of all players in one vector as follows: 
\begin{equation}
x = (h \;\; a),
\end{equation}
where $a$ denotes the concatenation of $(a_{1}\;\; \cdots\;\; a_{Q})$.

\subsection{Holistic feature stream}
\label{sec:holisticfeat}

This stream encodes a video segment $s$ in a spatio-temporal representation. We rely on state-of-the-art 3D CNNs over RGB frames. Training a 3D CNN from scratch requires thousands of videos \cite{karpathy2014large} that are not available for the proposed task. Recent work on deep neural networks for computer vision \cite{tran2015learning, feichtenhofer2016spatiotemporal, zeng2016title} shows that the activations of an upper layer of a CNN are useful for other related tasks without requiring fine-tuning. Thus, 3D CNN in which parameters are pre-trained with large-scale datasets can be used instead to leverage a huge amount of labeled training data \cite{jia2014caffe}. The proposed method utilizes a 3D CNN for action feature extraction pre-trained with a publicly available action recognition dataset, such as Sports-1M \cite{karpathy2014large}. Unlike our previous work \cite{tejerodepablos2016human}, which required to classify players' actions, it is not necessary to use a sport-specific action recognition dataset.%\tc{Another advantage of pre-trained CNN is that we do not need to adapt the input; they use normal RGB video segments. (why is this sentence necessary?)}

Two types of holistic representations of video segments extracted using 3D CNNs are employed, namely CNN-ISA \cite{le2011learning} and C3D \cite{tran2015learning}. Specifically, CNN-ISA provides a representation robust to local translation (e.g., small variations in players' or camera motion) while it is selective to frequency, rotation, and velocity of such motion. The details of CNN-ISA can be found in \cite{le2011learning}. CNN-ISA achieved state-of-the-art performance in well-known datasets for action recognition such as YouTube \cite{liu2009recognizing}, Hollywood2 \cite{marszalek2009actions}, and UCF sports \cite{rodriguez2008action}. Additionally, C3D features provide a representation of objects, scenes, and actions in a video. The network architecture and other details can be found in \cite{tran2015learning}. C3D pre-trained with the Sports-1M dataset achieved state-of-the-art performance on action recognition over the UCF101 dataset \cite{soomro2012ucf101}. 

This stream represents a video segment $s$ by using a holistic feature vector $y$ that corresponds to the output of one of the aforementioned 3D CNNs.

\subsection{Highlight classification using LSTM}

\begin{figure}[!t]
\centering
\includegraphics{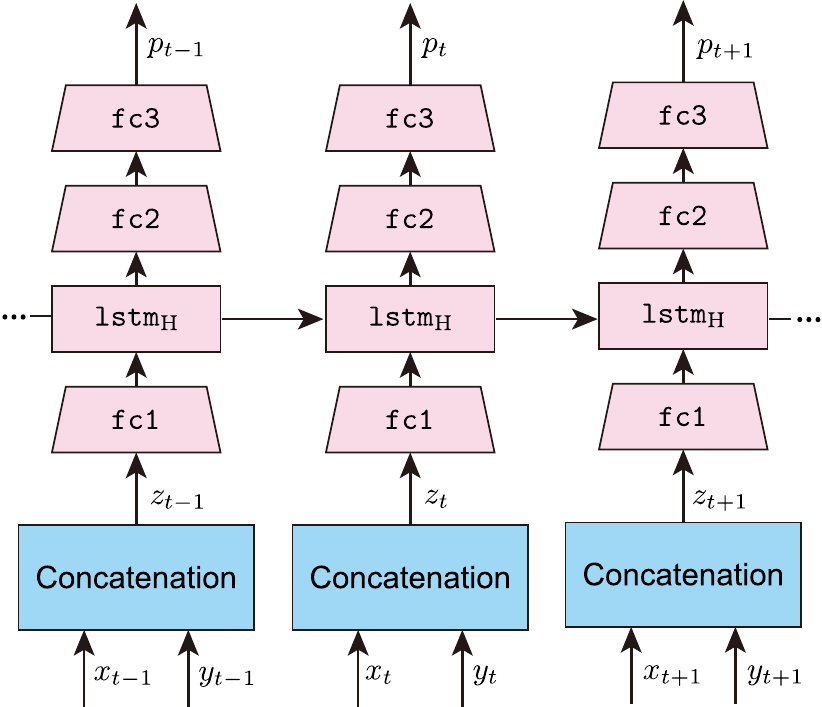}
\caption{The recurrent neural network architecture for highlight classification consists of a single LSTM layer and several fully-connected layers. The body joint-based features $x_t$ and holistic features $y_t$ extracted from video segment $s_t$ are input to calculate the probability $p_t$ that the segment is interesting.}
\label{fig:classification}
\end{figure}

Figure \ref{fig:classification} shows the network architecture designed to extract highlights of UGSV using the features $x_t$ and $y_t$ from video segment $s_t$. The temporal dependencies among video segments are modeled using an LSTM, and the network outputs the probability $p_t$ that the video segment $s_t$ is interesting. First, the features are concatenated to form vector $z_t = (x_t \;\; y_t)$. Vector $z_t$ then goes through a fully-connected layer to reduce its dimensionality.

It is assumed that interesting video segments are related to each other in time, in the same way a skillful boxer first feints a punch prior to hitting to generate an opening in the defense. Existing work in video summarization uses LSTMs to extract video highlights \cite{yang2015unsupervised} since it allows the modeling of temporal dependencies across longer time periods when compared to other methods \cite{yue2015beyond}. Following this concept, an LSTM layer is introduced to the network for highlight classification. The hidden state of the LSTM from each time step goes through two fully-connected layers, and this results in a final softmax activation of two units corresponding to ``interesting'' and ``uninteresting.''

\begin{figure}[!t]
\centering
\includegraphics[width=\columnwidth]{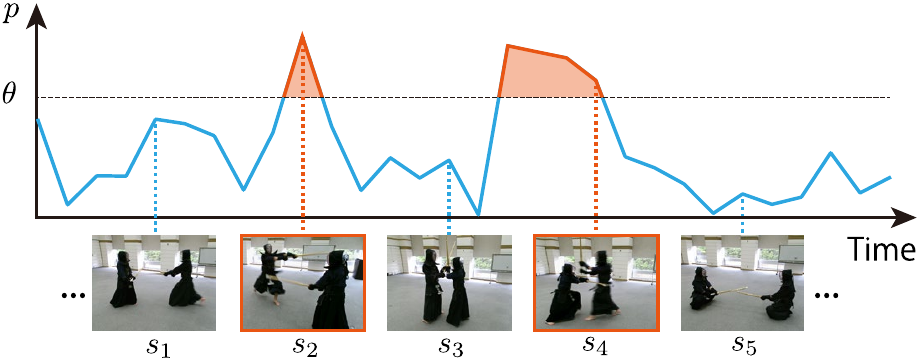}
%\vspace{-7mm}
\caption{A summary is generated by concatenating segments in which the probability $p_t \in [0,1]$ of being part of a highlight surpasses a certain threshold $\theta$. $\theta$ decreases from 1 until the desired summary length is reached.}
%\vspace{-4mm}
\label{fig:skimcurve}
\end{figure}

The proposed method provides control over the length $L$ of the output summary. The softmax activation of the unit corresponding to ``interesting'' is considered as the probability $p_t \in [0,1]$ that segment $s_t$ is part of a highlight, and skimming curve formulation \cite{truong2007video} is applied to the sequence of probabilities by decreasing a threshold $\theta$ from 1 until a set of segments whose total length is highest below $L$ is determined (Fig.~\ref{fig:skimcurve}). The segments in which the probability exceeds $\theta$ are concatenated to generate the output summary in the temporal order. Hence, the resulting summary may contain multiple consecutive interesting segments.

\subsection{Network training}

A pre-trained CNN is used in the holistic features stream (i.e., CNN-ISA or C3D) while the LSTMs and fully-connected layers are trained from scratch. Hence, during training, the parameters in the holistic feature stream (i.e., CNN layers) are fixed and those in the body joint-based feature stream (i.e., $\mathtt{lstm}_J$) and highlight classification (i.e., $\mathtt{fc1}$, $\mathtt{lstm}_H$, $\mathtt{fc2}$, and $\mathtt{fc3}$) are updated. 

The UGSV dataset contains video and ground truth labels $l_t \in \{0, 1\}$ for every second, where $l_t = 1$ implies that the period from $t$ sec to $t+1$ sec of the video is ``interesting'' and $l_t = 0$ otherwise. Label $l_t$ is assigned to its corresponding video segment $s_t$, which contains video second $t$ in its center. That is, with $\tau=3$, segment $s_t=\{t-1,t,t+1\}$ is assigned label $l_t$, segment $s_{t+1}=\{t,t+1,t+2\}$ is assigned label $l_{t+1}$, etc. (Fig.~\ref{fig:segmentation}).%and this covers the frames in $t-1$ to $t+2$ since $s_t$ captures the period from $t$ sec to $t_1$ sec in its center. 

With respect to training, cross-entropy loss $\ell$ is used:
\begin{equation}
\ell = \sum l_t \log p_t.
\end{equation}

\section{Experiments}
\label{sec:experiments}

The proposed method is evaluated objectively and subjectively. With respect to the objective evaluation, the performance of the proposed method is compared while using different representations of the players' actions. Specifically, only body joint features (3D or 2D), only holistic motion features (CNN-ISA or C3D), and a combination of both features are evaluated. Subsequently, the completeness of the highlights of the generated summaries are examined. With respect to the subjective evaluation, users with and without experience in the sport are surveyed to study their opinions with respect to the summaries.

\subsection{Implementation details}
\label{sec:imple}

For the evaluation, Kendo (Japanese fencing) was selected as an example of a sport. Kendo is a martial art featuring two players and a set of recognizable actions (e.g., attacking and parrying). We used the UGSV Kendo dataset in \cite{tejerodepablos2016human}, which contains 90 min of self-recorded Kendo matches divided in 10 RGB-D videos taken with a Microsoft Kinect v2, and extended it by adding 18 more self-recorded RGB-D Kendo videos. The total length of the videos is 246 min with a framerate of approximately 20 fps (since $\tau=3$~sec, $F=60$).

The body joint-based feature stream was configured for $Q=2$ players since Kendo is a two-player sport. The tracker in \cite{zhang2012microsoft} was used as is (without additional training) to estimate $J=15$ 3D body joint positions from depth maps: \textit{head}, \textit{neck}, \textit{torso}, \textit{right shoulder}, \textit{right elbow}, \textit{right wrist}, \textit{left shoulder}, \textit{left elbow}, \textit{left wrist}, \textit{right hip}, \textit{right knee}, \textit{right ankle}, \textit{left hip}, \textit{left knee}, and \textit{left ankle}. In order to estimate the 2D positions of the players' joints from the RGB frames, the CNN-based method proposed by Linna et al.~\cite{linna2016real} was used. We pre-trained this joint estimation CNN with the human pose dataset used by Linna et al.~\cite{linna2016real}, and then we fine-tuned it with our extended UGSV Kendo video dataset. The network provides $J=13$ joints (that is the same as the 3D case with the exception of \textit{neck} and \textit{torso}). Therefore, the size of vector $u^f_t$ is $Q\times J\times3=90$ in the case of 3D positions and $Q\times J\times2=52$ in the case of 2D. Given that the size of $\mathtt{lstm}_J$ is the same as that of the input and that the size of $a_t$ is $Q=2$, the feature vector $x_t$ for the stream is $\in \mathbb{R}^{92}$ for 3D and $\in \mathbb{R}^{54}$ for 2D.

\begin{table*}[t]
%\footnotesize
\begin{center}
\caption{Size of the learnable elements in the network with respect to the features used ($input\times output$).\newline Feature vector sizes are detailed in Section \ref{sec:imple})} \label{tab:netparam}
\begin{tabular}{ r|c c c|c c|c c }
  \hline
   & \multicolumn{3}{c|}{Body joint-based features only} & \multicolumn{2}{c|}{Holistic features only} & \multicolumn{2}{c}{Body joint-based and holistic features}\\
   & 3D joints & 2D joints & Action recognition & CNN-ISA & C3D & 3D joints + CNN-ISA & 2D joints + CNN-ISA\\
  \hline
  $\mathtt{lstm}_J$ &         $90\times90$  &         $52\times52$  &         ---  		&    		---  			&      ---  		&         $90\times90$	&      $52\times52$ \\
  $\mathtt{fc1}$ &         $92\times50$  &         $54\times50$  &         $402\times400$  &   $400\times400$  &  $4096\times400$  &  $492\times400$	& $454\times400$ \\
  $\mathtt{lstm}_H$ &  $50\times50$  &    $50\times50$  &     $400\times400$  &     $400\times400$  &   $400\times400$  &  $400\times400$  & $400\times400$ \\
  $\mathtt{fc2}$ &    $50\times20$   	&    $50\times20$	&         $400\times100$  &         $400\times100$  &   $400\times100$  &  $400\times100$  &  $400\times100$ \\
  $\mathtt{fc3}$ &         $20\times2$  &         $20\times$2  &         	$100\times2$  &         $100\times2$  &         $100\times2$  &         $100\times2$  &  $100\times2$ \\
  \hline
\end{tabular}
\end{center}
%\vspace{-6mm}
\end{table*}

With respect to the holistic feature stream, either the CNN-ISA \cite{le2011learning} or C3D \cite{tran2015learning} networks were used. The UGSV Kendo dataset is not sufficiently large to train the CNNs from scratch, and thus networks pre-trained with an action recognition dataset were used. The CNN-ISA was trained in an unsupervised way with the Hollywood2 dataset that consists of 2859 videos \cite{marszalek2009actions}. For this network, we followed the configuration in \cite{wang2009evaluation}. We used a vector quantization representation of the extracted features with a codebook size of 400, thereby resulting in a feature vector $y_t \in \mathbb{R}^{400}$ for each segment $s_t$. The C3D was trained with the Sports-1M dataset \cite{karpathy2014large} that consisted of 1.1 million videos of sports activities. The C3D features were extracted as indicated in \cite{tran2015learning} by uniformly sub-sampling 16 frames out of approximately 60 frames in $s_t$ (the number of frames in $s_t$ may vary for different segments due to the variable framerate of Microsoft Kinect v2) and subsequently the activations from layer \texttt{fc6} (i.e., $y_t \in \mathbb{R}^{4096}$) were extracted.

The proposed method was implemented in Chainer \cite{tokui2015chainer} running on Ubuntu Trusty (64 bit), installed in a computer with an Intel Core i7 processor and 32GB of RAM, and a GeForce GTX TITAN X graphics card. In average, it roughly took 300 min to train the network until convergence over our Kendo dataset. For testing, the average processing time of a video is approximately 5 sec (see Table \ref{tab:dataset} for video durations). The learning rate was calculated by the adaptive moment estimation algorithm (Adam) \cite{kingma2015adam} with $\alpha = 0.001$. Sigmoid activation was introduced after the fully-connected layers. Table \ref{tab:netparam} summarizes the number of learnable parameters for each layer, which varies based on the choice of features.

%Since recording the dataset and annotating the videos was a laborious task, we only applied our method to Kendo UGSV. However, we are firmly convinced that these results can be extrapolated to similar sports like boxing, fencing, and basically to any sport where players' actions can be tracked.

\subsection{Results}

For annotating the ground truth, 15 participants were invited and divided into two groups, namely experienced (\textit{E}, 5 people) and inexperienced (\textit{NE}, 10 people), based on their experience in the target sport (i.e., Kendo). It was assumed that the highlights preferred by the \textit{E} and \textit{NE} groups would exhibit significant variations, and an aim of the study included evaluating the extent to which the proposed method adapts to the needs of each group. For this, the participants annotated manually the highlights of the 28 videos. The ground truth labels of the videos were separately obtained for both E and NE groups. With respect to each one-second period $t$ of video, the ground truth label is $l_t=1$ if at least 40\% of the participants annotated it as interesting (i.e., 2 people in group \textit{E} and 4 people in group \textit{NE}). Otherwise, $l_t=0$. Due to group \textit{E}'s technical knowledge of Kendo, their highlights contain very specific actions (e.g., decisive strikes and counterattacks). Conversely, group \textit{NE} selected strikes as well as more general actions (e.g., parries and feints), and thus their labeled highlights are almost three times as long as group \textit{E}'s (please refer to the durations in Appendix \ref{sec:appendixa}). 

The network was separately trained with each group's ground truth labels in the leave-one-out (LOO) fashion, i.e., 27 videos were used for training and a summary of the remaining video was generated for evaluation purposes. The CNN for 2D pose estimation was trained independently prior to each experiment, it was fine-tuned with the 27 training videos in order to estimate the joints of the video used for evaluation. This process was repeated for each video and for each group \textit{E} and \textit{NE}, to result in 28 experienced summaries and 28 inexperienced summaries. The generated summaries had the same length $L$ as their respective ground truth for a fair comparison. Figure \ref{fig:skimcurve} illustrates a few examples of the frames of a video as well as highlight frames extracted by the proposed method (framed in orange).

\subsubsection{Objective evaluation by segment f-score}
\label{sec:interes}

The ability of the proposed method to extract highlights was evaluated in terms of the f-score. In the proposed method, a one-second period of video is as follows:
\begin{itemize}
\item true positive (TP), if it is in the summary and $l_t = 1$,
\item false positive (FP), if it is in the summary but $l_t = 0$,
\item false negative (FN), if it is not in the summary but $l_t = 1$,% or
\item true negative (TN), if it is not in the summary and $l_t = 0$.
\end{itemize}
The f-score is subsequently defined as follows:
\begin{equation}
\textrm{f-score}=\frac{2\textrm{TP}}{2\textrm{TP}+\textrm{FP}+\textrm{FN}}
.
\end{equation}
%and the accuracy as
%\begin{equation}
%ACC=\frac{TP + TN}{TP + FP + FN + TN}
%\end{equation}

Table \ref{tab:objecfeat} shows the f-scores for the summaries generated with the labels of both \textit{E} and \textit{NE} groups. In addition to the features described in Section \ref{sec:imple}, it includes the results of using the features from our previous work in UGSV summarization \cite{tejerodepablos2016human}. The features were obtained by feeding the 3D body joint representation of players' actions to the action recognition method in \cite{tejerodepablos2016flexible} and considering the action classification results. Additionally, the proposed architecture was also compared with that of the method used in the previous work \cite{tejerodepablos2016human} that uses a hidden Markov model with Gaussian mixture emission (GMM-HMM) over the same action recognition results mentioned above. Finally, the results of using $k$-means clustering are included, since $k$-means is widely accepted as a baseline for user-generated video summarization \cite{cong2012towards}. To implement the $k$-means clustering baseline, the video segments $S$ were clustered based on the concatenated features \textit{3D joints + CNN-ISA}, and the summary was created by concatenating in time the cluster centroids. The number of clusters for each video were configured such that the resulting summary length is equal to that of the ground truth.
%UNCOMMENT
%\begin{table}[!h]
%\begin{center}
%\caption{Objective evaluation of our summarization method.} \label{tab:quant}
%\begin{tabular}{ r|c c|c c }
%  \hline
%   Extracted features  & \multicolumn{2}{c|}{Annotations \textit{E}} & \multicolumn{2}{c}{Annotations \textit{NE}}\\
%   & F1 & ACC & F1 & ACC\\
%  \hline
%  3D joints &         0.53  &         X  &         0.83  &         X \\
%  2D joints &         0.44  &         X  &         0.77  &         X \\
%  HAR &		0.48 &		 X & 		0.76 &		 X\\
%  \hline
%  CNN-ISA &   0.49  &         X  &         0.79  &         X \\
%  C3D &         0.27  &         X  &         0.60  &         X \\
%  \hline
%  \textbf{3D joints + CNN-ISA} &   \textbf{0.58}  & \textbf{X}  & \textbf{0.85}  &  \textbf{X} \\
%  2D joints + CNN-ISA &         0.57  &         X  &         0.81  &         X \\
%  \hline
%  GMM-HMM &         0.44  &         X  &         0.79  &         X \\
%  Clustering &         ?  &         X  &         ?  &         X \\
%  Evenly spaced &         ?  &         X  &         ?  &         X \\
%  \hline
%\end{tabular}
%\end{center}
%\end{table}
\begin{table}[!t]
\begin{center}
\caption{F-score comparison of different combinations of features and other UGSV summarization methods.} \label{tab:objecfeat}
\begin{tabular}{ r|r|c|c }
  \hline
   \multicolumn{2}{c|}{Method}  & Group \textit{E} & Group \textit{NE} \\ \hline
  \multirow{3}{*}{\parbox{2.3cm}{Body joint-based\newline features}} & 3D joints &       0.53  &         0.83  \\
   & 2D joints &         0.45  &       0.77  \\
   & Action recognition \cite{tejerodepablos2016human} &		0.48 & 		0.76 \\
   \hline
  \multirow{2}{*}{\parbox{2.3cm}{Holistic features}} & CNN-ISA &   0.50  &         0.79  \\
   & C3D &         0.27  &         0.60   \\
  \hline
  \multirow{2}{*}{\parbox{2.3cm}{Body joint-based\newline and holistic features}} & 3D joints + CNN-ISA &   \textbf{0.58}  & \textbf{0.85}  \\
  & 2D joints + CNN-ISA &         0.57  &         0.81  \\
  \hline
  \hline
  \multirow{2}{*}{\parbox{2.3cm}{Other UGSV\newline summarization}} & $k$-means clustering &	0.28	&	0.61	\\
  \cline{2-4}
  & Without $\mathtt{lstm}_H$		&         0.48  &         0.8  \\
  \cline{2-4}
  & GoogLeNet and BiLSTM \cite{zhang2016video}	&         0.27  &       0.65  \\
  \cline{2-4}
  & GMM-HMM \cite{tejerodepablos2016human}		&         0.44  &         0.79  \\  
  \hline
\end{tabular}
\end{center}
\end{table}

With respect to using a single feature (i.e.~3D joins, 2D joints, CNN-ISA, C3D, or action recognition), 3D joints obtain the best performance. Although C3D features perform well in action recognition in heterogeneous video \cite{tran2015learning}, the results were worse than that of other features in our summarization task. The dimensionality of the C3D features (4096) is significantly higher when compared to that of others, and thus our dataset may not be sufficient to train the network well. Fine-tuning C3D using the Kendo dataset might improve its performance. In contrast, CNN-ISA also uses RGB frames and obtains better results when compared to those of C3D, most likely due to the lower dimensionality of its features (400). This implies that it is also possible to obtain features from RGB frames that allow the modeling of UGSV highlights. The decrease in the performance of 2D joints with respect to 3D joints may indicate that view variations in the same pose negatively affect the body joint-based features stream. The action recognition feature had an intermediate performance. A potential reason is that the action recognition feature is based on a classic approach for classification, so useful cues contained in the 3D body joint positions degenerated in this process. From these results, the features that performed better for highlight classification correspond to CNN-ISA holistic features and 3D body joint-based features. 

Several state-of-the-art action recognition methods enjoy improvements in performance by combining handcrafted spatio-temporal features (e.g., dense trajectories) and those learned via CNNs \cite{tran2015learning, feichtenhofer2016spatiotemporal}. This is also true in the present work where a combination of CNN-ISA with 3D joints achieves the best performance. The combination of CNN-ISA with 2D joints also provides a considerable boost in performance and especially for the experienced summaries. This supports our hypothesis that a two-streams architecture also provides better results for UGSV summarization.% We can also confirm the usefulness of 3D body joint-based representation of actions in order to model our highlights. However, we do not rule out the possibility that, with a sufficiently large dataset of Kendo videos, we could fine-tune our holistic stream to obtain a better performance.

Finally, the lowest part of Table \ref{tab:objecfeat} shows the results of other summarization methods. The results of the proposed method outperform the results of previous works, as well as those of the clustering-based baseline. While clustering allows a wider variety of scenes in the summary, this is not a good strategy for UGSV summarization that follows a different criterion based on interestingness.
To investigate the necessity of capturing temporal dependencies of our action features (3D joints + CNN-ISA), we replaced $\mathtt{lstm}_H$ in our network with a fully-connected layer of the same size (400$\times$400). This experiment allowed us to draw some interesting conclusions: Modeling the temporal relationship among sequential action features allows for an improved performance. Moreover, this improvement is more noticeable in the case of experienced users, because of their more elaborated labeling of interesting actions.
Then, we compared the proposed method with the state-of-the-art summarization method of Zhang et al. \cite{zhang2016video}. Zhang et al. extract features from each frame with a pre-trained CNN for image recognition (i.e. GoogLeNet \cite{szegedy2015going}) and feeds those features to a bidirectional LSTM to model the likelihood of whether the frames should be included in the summary. As shown in Table \ref{tab:objecfeat}, in spite of the more sophisticated temporal modeling in \cite{zhang2016video}, the performance is lower than most feature combinations in our method. This is most likely due to the particularities of sports video; GoogLeNet, as a network pre-trained for image classification, may not be able to extract features that represent different actions of a sport. Moreover, whereas our features are extracted from a video segment (which contains several frames), features in \cite{zhang2016video} are extracted from a single frame, and thus cannot represent continuous motion.
The proposed method also outperforms our previous work \cite{tejerodepablos2016human}, which used the classification results of an action recognition method to train a GMM-HMM for highlight modeling. We can conclude that it is not necessary to explicitly recognize players' actions for UGSV summarization, which may actually degrade performance when compared to that in the case of directly using action recognition features. 

% Add accuracy results: ACC = (TP + TN) / (TP + FP + FN + TN)

\subsubsection{Objective evaluation by highlight completeness}
\label{sec:complet}

\begin{figure}[!t]
\centering
\includegraphics[scale=.5]{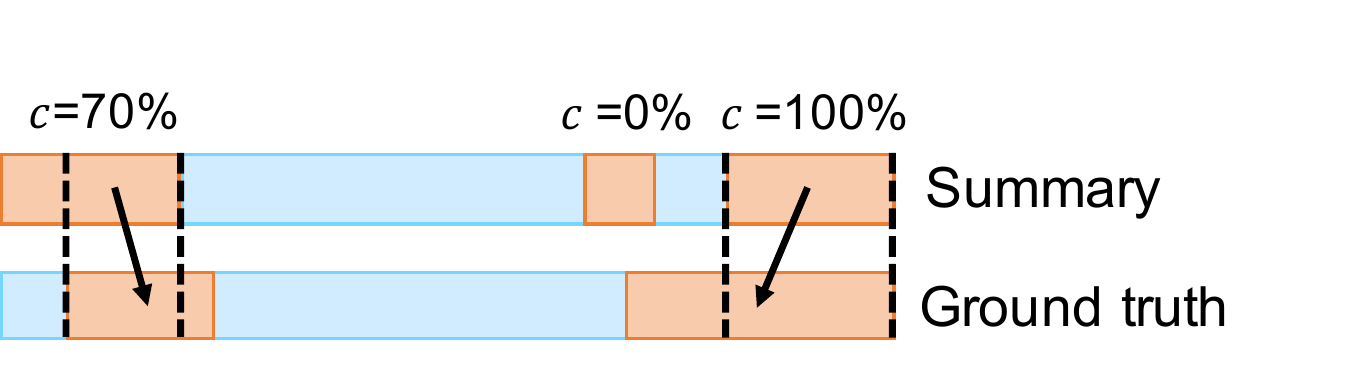}
\caption{Association of highlights with respect to the greedy algorithm. Each highlight in the ground truth is uniquely associated to a highlight in the generated summary (two summary highlights cannot share the same ground truth highlight). The completeness of a summary highlight corresponds to the percentage of overlap with the ground truth (0\% if unassociated).}
\label{fig:comple}
\end{figure}

\begin{figure}[!t]
\centering
\includegraphics[scale=.5]{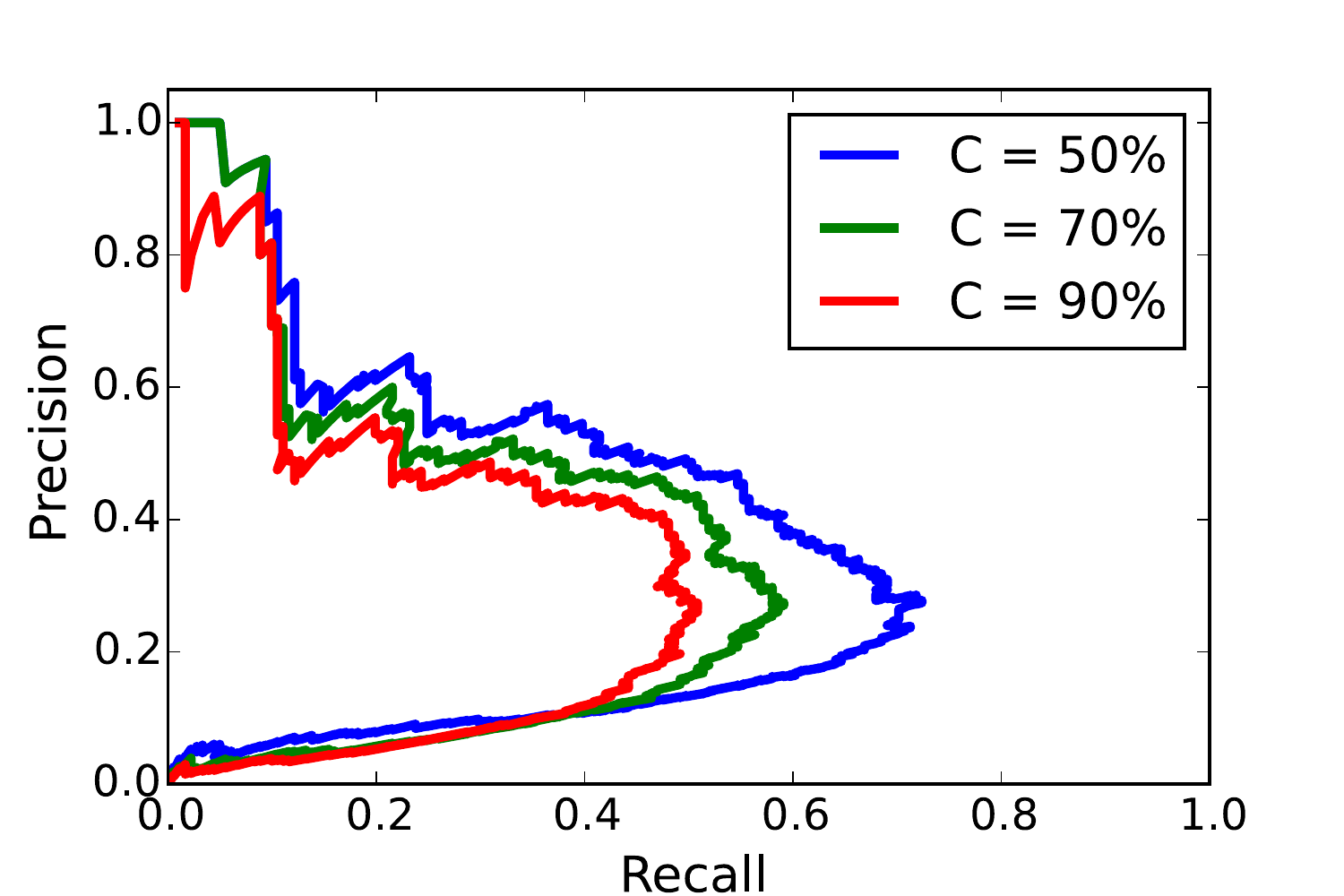}
\includegraphics[scale=.5]{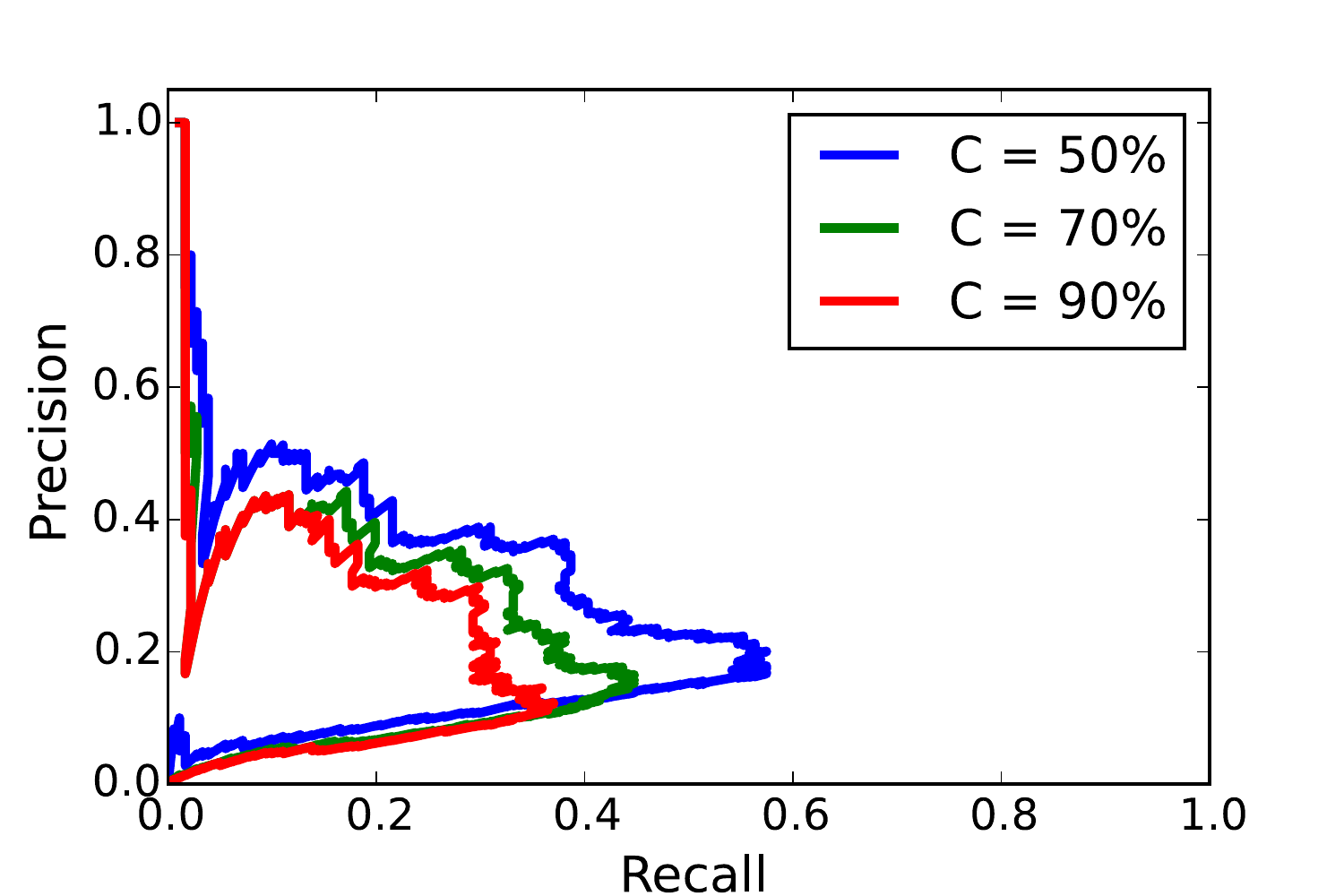}
\caption{Recall-precision curves for different completeness values (up: labels \textit{E}, down: labels \textit{NE}). The gap between the curves $C=50\%$ and $C=70\%$ shows that a significant number of the highlights are missing for a maximum of half the interesting segments. As $\theta$ varies, the appearing of incomplete highlights affects the association of highlights-ground truth, resulting in a jagged curve.}
\label{fig:rec-pre}
\end{figure}

\begin{figure*}[!thp]
\centering
\includegraphics[scale=.5]{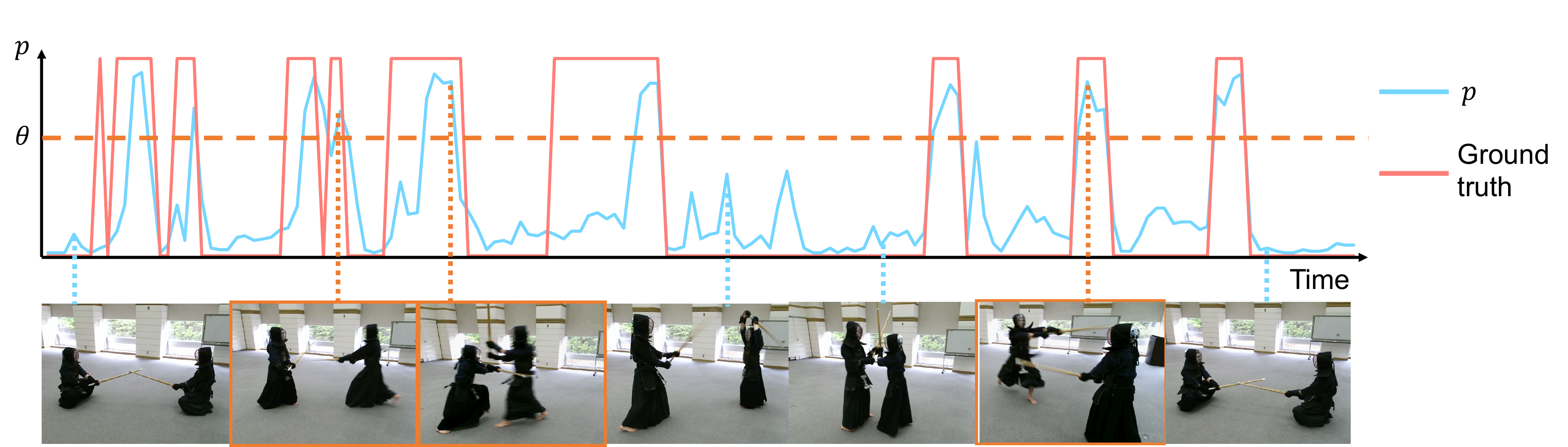}
\caption{Original length: 10 min 40 sec. Summary length: 1 min. The highlights summary is generated by applying a threshold $\theta$ to the probability of interestingness $p$. Video segments with higher $p$ are extracted prior to segments with lower $p$, and thus in a few cases the beginning/end segments of the highlights are missing when compared to the ground truth.}
\label{fig:threshold}
\end{figure*}

A highlight may consist of consecutive video segments. Hence, although missing a segment may not significantly impact the f-score, it affects the continuity of the video, and thereby the comprehensibility and the user experience of the summary. Given this, a criterion is defined to evaluate the completeness $c$ of an extracted highlight as the fraction of overlap between the extracted highlight and its associated ground truth highlight. The association between extracted and ground truth highlights is not trivial, and it was performed by using a greedy algorithm in which the total $c$ of all highlights is maximized (Fig. \ref{fig:comple}). An extracted highlight is considered as a TP if its completeness $c$ exceeds a certain percentage $C$\%, and based on this, the precision and recall of the highlights are calculated as follows:
\begin{equation}
\textrm{precision}=\frac{\textrm{TP}}{\textrm{TP}+\textrm{FP}}
,
\;\;\;\;\;\;
\textrm{recall}=\frac{\textrm{TP}}{\textrm{TP}+\textrm{FN}}
.
\end{equation}
In the experiment, the threshold $\theta$ varies from 0 to 1 over the probability $p$ to generate the recall-precision curve of group \textit{E} and \textit{NE}.

Figure \ref{fig:rec-pre} shows the curves produced for $C = $ 50\%, 70\%, and 90\%. We observe that reducing $C$ to 50$\%$ significantly increases the number of complete highlights. The presence of incomplete highlights is attributed to the way highlights are extracted. First, the \textit{high $p$ segments} are extracted, and then the highlight is completed with \textit{low $p$ segments} as the threshold $\theta$ decreases (Fig. \ref{fig:threshold}). However, prior to the completion of a highlight, \textit{high $p$ segments} from other highlights are extracted and, in a few cases, the 
\textit{low $p$ segments} are never extracted. Specifically, the parts before and after an interesting Kendo technique normally correspond to \textit{low $p$ segments} since they are not present in every ground truth highlight annotated by the participants.

The reason for the increased number of incomplete segments (less TP) in the \textit{NE} summaries is because the inexperienced group annotated a higher number of highlights.

\subsubsection{Subjective evaluation}

The same participants who annotated the original videos were asked to participate in a survey to assess their opinion on the ground truth and the generated summaries. The three videos with the highest, median and lowest f-scores (averaged over groups $E$ and $NE$) were selected. With respect to each video, participants were shown the ground truth and the summaries generated with the best feature combination (i.e., \textit{3D joints + CNN-ISA}) using both group \textit{E} and \textit{NE} labels. As a result, each participant watched 12 videos (3 f-scores $\times$ 4 video types).

The participants were asked to:
\begin{itemize}
\item (Q1) assign a score in a Likert scale from 1 (very few highlights are interesting) to 5 (most highlights are interesting) based on their satisfaction with the contents of each of the 12 videos.
\item (Q2) state their opinion on the videos and the criteria followed while assigning a score.
\end{itemize}
Table \ref{tab:subjec} shows the results of Q1 grouped by video type and video f-score. The scores are averaged for group \textit{E} and \textit{NE} separately.

\begin{table*}[t]
\begin{center}
\caption{Subjective evaluation results with respect to the video type and f-score.\newline Each cell contains the mean $\pm$ the standard deviation of the scores (from 1 to 5).} \label{tab:subjec}
\begin{tabular}{c|cccc|ccc}
  \hline
   \multirow{2}{*}{} & \multicolumn{4}{c|}{Video type} & \multicolumn{3}{c}{Video f-score} \\
    & Ground truth \textit{E} & Ground truth \textit{NE} & Summary \textit{E} & Summary \textit{NE} & Highest & Median & Lowest \\
  \hline
  Group \textit{E} & 3.2$\pm$0.99 & 3.07$\pm$1.04 & 2.6$\pm$1.23 & 2.73$\pm$0.87 & 3.3$\pm$0.95 & 2.85$\pm$0.97 & 2.55$\pm$1.18 \\
  Group \textit{NE} & 3.57$\pm$0.72 & 3.5$\pm$1.07 & 3.2$\pm$0.83 & 2.9$\pm$0.97 & 3.48$\pm$0.83 & 3.03$\pm$0.91 & 3.38$\pm$0.95 \\
  \hline
\end{tabular}
\end{center}
\end{table*}

% ABOUT THE SUBJECTIVE EVALUATION: Even if our method did not extract all the highlights selected by group E, the summaries generated with the experienced annotations got the best scores for all questions in the survey by both groups.

In the context of Q1, with respect to the video type, both experienced and inexperienced participants assigned a higher score to the ground truth videos than to the generated summaries. This is because some of the summaries contain uninteresting video segments and also the completeness of the highlights is worse when compared to that of ground truth videos. The potential reasons as to why the ground truth videos did not obtain a perfect score are mainly attributed to the following two factors: (1) The ground truth summaries are created by combining labels from several participants via majority voting, and thus the original labels of each participant are lost. (2) The ground truth also contains incomplete highlights due to errors when the participants annotated the videos. Additionally, experienced participants preferred the \textit{NE} ground truth to the \textit{E} summaries plausibly because they do not find incomplete highlights interesting since context is missing. Conversely, inexperienced participants tend to appreciate the highlights from the experienced participants more than their own highlights. This is potentially because the highlights from the experienced participants are briefer and contain certain techniques (e.g. counterattacks) that make summaries more interesting when compared to those of the inexperienced participants. The results for Q1 in terms of the f-score type demonstrate the high correlation to the f-score (i.e., a video with a higher f-score tends to receive a higher subjective score).

With respect to Q2, participants provided their opinion on the summaries.  A few experienced participants found the highlights as too short and this even included complete highlights in the ground truth. This occurs because only the segments labeled as highlights by at least 40\% of the participants (i.e., 2 people in group \textit{E} and 4 people in group \textit{NE}) were included in the ground truth, and thus some labeled segments were left out. Inexperienced participants state the usefulness of the proposed method to extract highlights based on interesting actions as well as time saved by watching the highlights as opposed to the whole video. In addition, for a few inexperienced participants incomplete highlights make the summaries difficult to follow.% Some inexperienced users suggested applying slow motion replays on the most salient video segments.

From this evaluation, we conclude that the labels from experienced users contain a better selection of Kendo techniques. Due to the negative impact of incomplete highlights on the summaries, it is necessary to consider extra temporal consistency in $p_t$. One possibility is to replace the skimming curve formulation-based highlight extraction with an algorithm that takes into account the completeness of the highlights. Also, another possibility is not to combine the labels of several participants since it introduces incomplete highlights (Section \ref{sec:complet}) and alters personal preferences. Thus, instead of combining labels from different participants, another possibility is to create personalized summaries with a higher quality ground truth or to include user profiles such as that proposed in \cite{nitta2009automatic}.

\section{Conclusion}
\label{sec:conclusions}

This paper has described a novel method for automatic summarization of UGSV, especially demonstrating the results for Kendo (Japanese fencing) videos. Given the lack of editing conventions that permit the use of heuristics, a different cue, i.e., players' actions, is used to acquire high-level semantics from videos to generate a summary of highlights. The presented two-stream method combines body joint-based features and holistic features for highlights extraction. The best combination among the evaluated features corresponds to a combination of 3D body joint-based features and CNN-ISA features \cite{le2011learning}). In contrast to the previous work \cite{tejerodepablos2016human}, the results indicate that it is not necessary to explicitly recognize players' actions in order to determine highlights. Alternatively, deep neural networks are leveraged to extract a feature representation of players' actions and to model their temporal dependency. Specifically, LSTM is useful to model the temporal dependencies of the joint positions of players' bodies in each video segment as well as the highlights in the entire video. In order to generate appealing summaries, players' 3D body joint positions from depth maps offer the best performance. However, in the absence of depth maps, 2D body joint positions and holistic features extracted from RGB images are also used for summarization.

The future work includes improving the architecture of the network and fine-tuning it in the end-to-end manner with a larger dataset to illustrate its potential performance. It also includes evaluating the method in the context of a wider variety of sports (e.g., boxing, fencing, and table tennis).

% use section* for acknowledgment
\section*{Acknowledgment}

This work was supported in part by JSPS KAKENHI Grant Number 16K16086.

% if have a single appendix:
%\appendix[Proof of the Zonklar Equations]
% or
%\appendix  % for no appendix heading
% do not use \section anymore after \appendix, only \section*
% is possibly needed

% use appendices with more than one appendix
% then use \section to start each appendix
% you must declare a \section before using any
% \subsection or using \label (\appendices by itself
% starts a section numbered zero.)
%

\appendices
\section{}
\label{sec:appendixa}

Table \ref{tab:dataset} lists the duration of the videos in the dataset used in the experiments and their respective ground truth highlights as annotated by users.

\begin{table}[!ht]
%\footnotesize
\begin{center}
\caption{Duration of the video dataset and ground truths.} \label{tab:dataset}
\begin{tabular}{ r|c|c|c }
  \hline
  ID & Original video & Ground truth \textit{E} & Ground truth \textit{NE} \\
  \hline
  \#1	&	10 min 48 sec	&	1 min 11 sec	&	2 min 21 sec \\
  \#2	&	5 min 10 sec	&	49 sec	&	1 min 7 sec \\
  \#3	&	5 min 18 sec	&	1 min 9 sec	&	1 min 58 sec \\
  \#4	&	9 min 37 sec	&	1 min 37 sec	&	2 min 17 sec \\
  \#5	&	9 min 59 sec	&	2 min 33 sec	&	2 min 42 sec \\
  \#6	&	10 min 5 sec	&	1 min 28 sec	&	2 min 55 sec \\
  \#7	&	10 min 3 sec	&	48 sec	&	1 min 45 sec \\
  \#8	&	10 min 10 sec	&	45 sec	&	2 min 14 sec \\
  \#9	&	5 min 17 sec	&	32 sec	&	1 min 14 sec \\
  \#10	&	5 min 14 sec	&	22 sec	&	1 min 30 sec \\
  \#11	&	4 min 58 sec	&	53 sec	&	1 min 50 sec \\
  \#12	&	20 min 40 sec	&	1 min 24 sec	&	4 min 14 sec \\
  \#13	&	10 min 15 sec	&	53 sec	&	2 min 50 sec \\
  \#14	&	10 min 16 sec	&	58 sec	&	5 min 8 sec \\
  \#15	&	10 min 37 sec	&	47 sec	&	2 min 44 sec \\
  \#16	&	10 min 37 sec	&	34 sec	&	2 min 21 sec \\
  \#17	&	5 min 14 sec	&	16 sec	&	1 min 44 sec \\
  \#18	&	5 min 4 sec	&	32 sec	&	2 min 21 sec \\
  \#19	&	10 min 57 sec	&	38 sec	&	2 min 11 sec \\
  \#20	&	5 min 36 sec	&	27 sec	&	1 min 21 sec \\
  \#21	&	5 min 36 sec	&	33 sec	&	1 min 35 sec \\
  \#22	&	10 min 48 sec	&	58 sec	&	1 min 59 sec \\
  \#23	&	9 min 44 sec	&	1 min 11 sec	&	2 min 48 sec \\
  \#24	&	10 min 23 sec	&	54 sec	&	2 min 25 sec \\
  \#25	&	10 min 7 sec	&	28 sec	&	1 min 57 sec \\
  \#26	&	10 min 40 sec	&	49 sec	&	2 min 5 sec \\
  \#27	&	4 min 59 sec	&	33 sec	&	2 min 13 sec \\
  \#28	&	8 min 13 sec	&	47 sec	&	2 min 10 sec \\
  \hline
  Total	&	4 hours 6 min 11 sec	&	24 min 49 sec	&	1 hour 3 min 59 sec \\
\end{tabular}
\end{center}
%\vspace{-6mm}
\end{table}

% you can choose not to have a title for an appendix
% if you want by leaving the argument blank
%\section{Equations}
%Appendix two text goes here.

% Can use something like this to put references on a page
% by themselves when using endfloat and the captionsoff option.
\ifCLASSOPTIONcaptionsoff
  \newpage
\fi

% trigger a \newpage just before the given reference
% number - used to balance the columns on the last page
% adjust value as needed - may need to be readjusted if
% the document is modified later
%\IEEEtriggeratref{8}
% The "triggered" command can be changed if desired:
%\IEEEtriggercmd{\enlargethispage{-5in}}

% references section

% can use a bibliography generated by BibTeX as a .bbl file
% BibTeX documentation can be easily obtained at:
% http://mirror.ctan.org/biblio/bibtex/contrib/doc/
% The IEEEtran BibTeX style support page is at:
% http://www.michaelshell.org/tex/ieeetran/bibtex/
% argument is your BibTeX string definitions and bibliography database(s)

\bibliographystyle{IEEEtran}
\bibliography{ieee-transMM}

% <OR> manually copy in the resultant .bbl file
% set second argument of \begin to the number of references
% (used to reserve space for the reference number labels box)

%\begin{thebibliography}{1}
%\bibitem{IEEEhowto:kopka}
%H.~Kopka and P.~W. Daly, \emph{A Guide to \LaTeX}, 3rd~ed.\hskip 1em plus 0.5em minus 0.4em\relax Harlow, England: Addison-Wesley, 1999.
%\end{thebibliography}

% biography section
% 
% If you have an EPS/PDF photo (graphicx package needed) extra braces are
% needed around the contents of the optional argument to biography to prevent
% the LaTeX parser from getting confused when it sees the complicated
% \includegraphics command within an optional argument. (You could create
% your own custom macro containing the \includegraphics command to make things
% simpler here.)
%\begin{IEEEbiography}[{\includegraphics[width=1in,height=1.25in,clip,keepaspectratio]{mshell}}]{Michael Shell}
% or if you just want to reserve a space for a photo:

\vspace{-1cm}

\begin{IEEEbiography}
[{\includegraphics[width=1in,height=1.25in,clip,keepaspectratio]{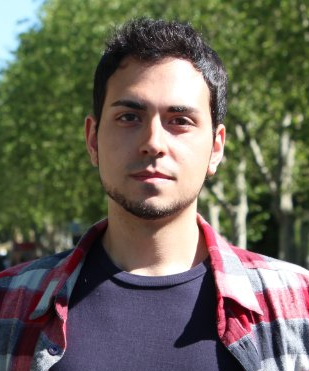}}]
{Antonio Tejero-de-Pablos}
received a B.E. in telematics engineering and a M.E.in telecommunications engineering from University of Valladolid, Spain, in 2009 and 2012 respectively. He received a Ph.D. in information sciences from Nara Institute of Science and Technology (NAIST), Japan, in 2017. He is currently a researcher at the University of Tokyo. His research interests include action recognition, video summarization, user interfaces and general-purpose computing on GPU. He is a member of the IEEE.
\end{IEEEbiography}

\vspace{-1cm}

\begin{IEEEbiography}
[{\includegraphics[width=1in,height=1.25in,clip,keepaspectratio]{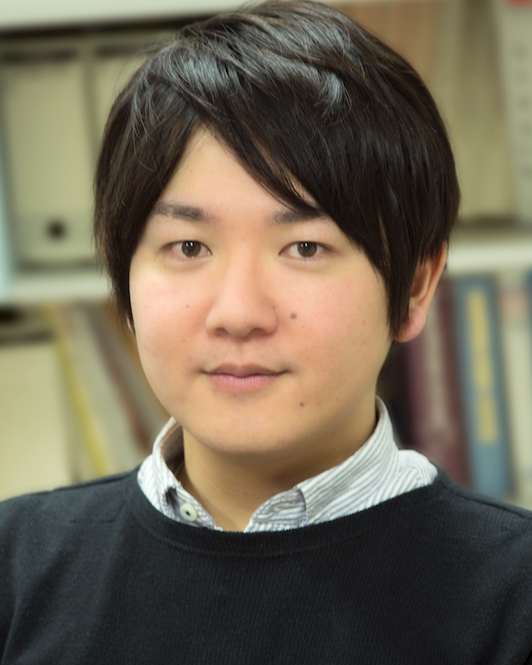}}]
{Yuta Nakashima}
received a B.E. and a M.E. in communication engineering and a Ph.D. in engineering from Osaka University, Osaka, Japan in 2006, 2008, and 2012, respectively. He was an assistant professor at Nara Institute of Science and Technology from 2012 to 2016 and is currently an associate professor at the Institute for Datability Science, Osaka University. He was a visiting scholar at UNCC in 2012 and at CMU from 2015-2016. His research interests include computer vision and machine learning and their applications. His main research focus includes video content analysis using machine learning approaches. He is a member of IEEE, ACM, IEICE, and IPSJ.
\end{IEEEbiography}

\vspace{-1cm}

\begin{IEEEbiography}
[{\includegraphics[width=1in,height=1.4in,clip]{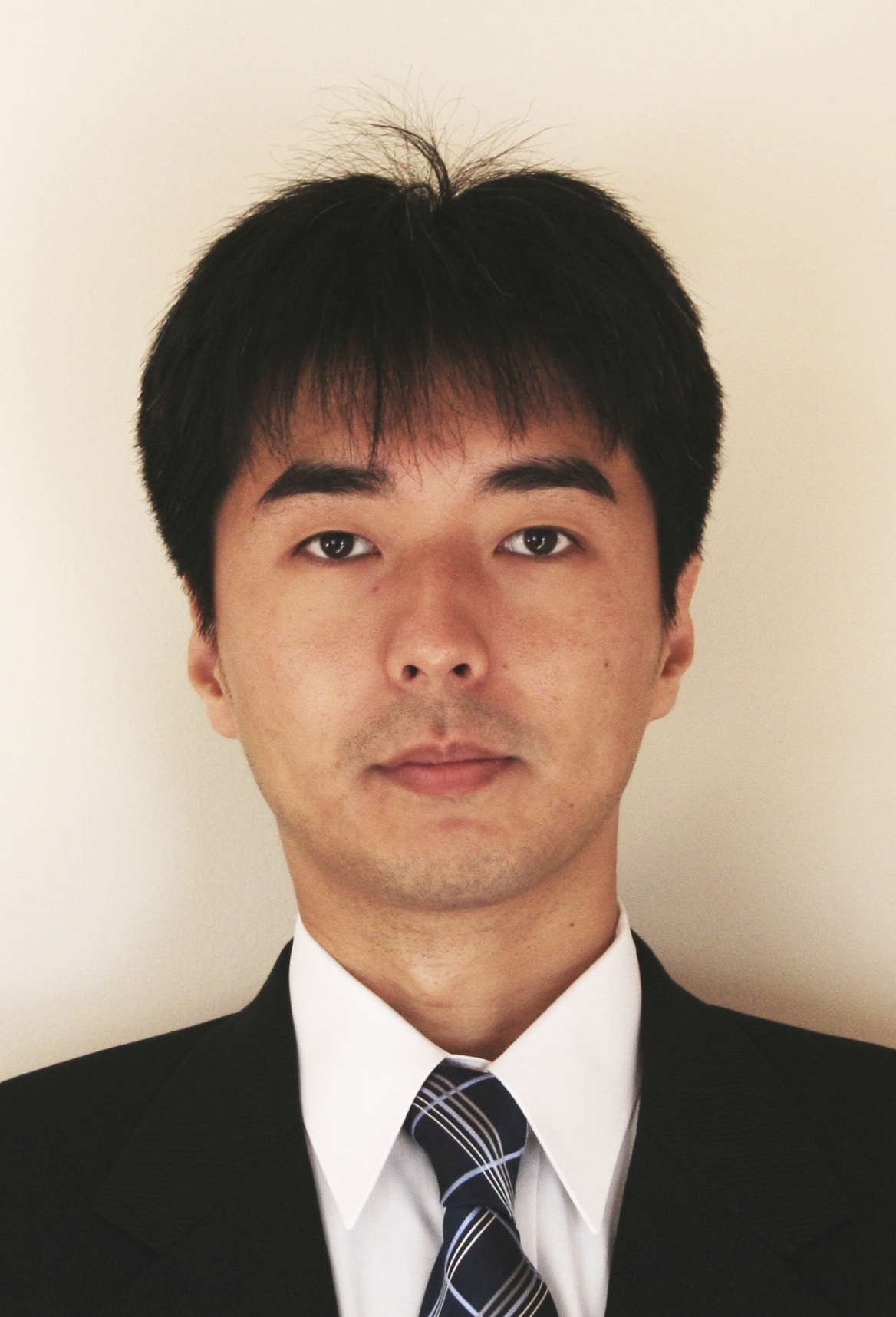}}]
{Tomokazu Sato}
received a B.E. in computer and system science from Osaka Prefecture University in 1999. He received a M.E. and a Ph.D. in information sciences from Nara Institute of Science and Technology (NAIST) in 2001 and 2003, respectively. He was an assistant
professor at NAIST during 2003-2011. He served as a visiting researcher at Czech Technical University in Prague from 2010-2011. He is an associate professor at NAIST since 2011. His research interests include computer vision and mixed reality.
\end{IEEEbiography}

%\vspace{-1cm}

\begin{IEEEbiography}
[{\includegraphics[width=1in,height=1.25in,clip,keepaspectratio]{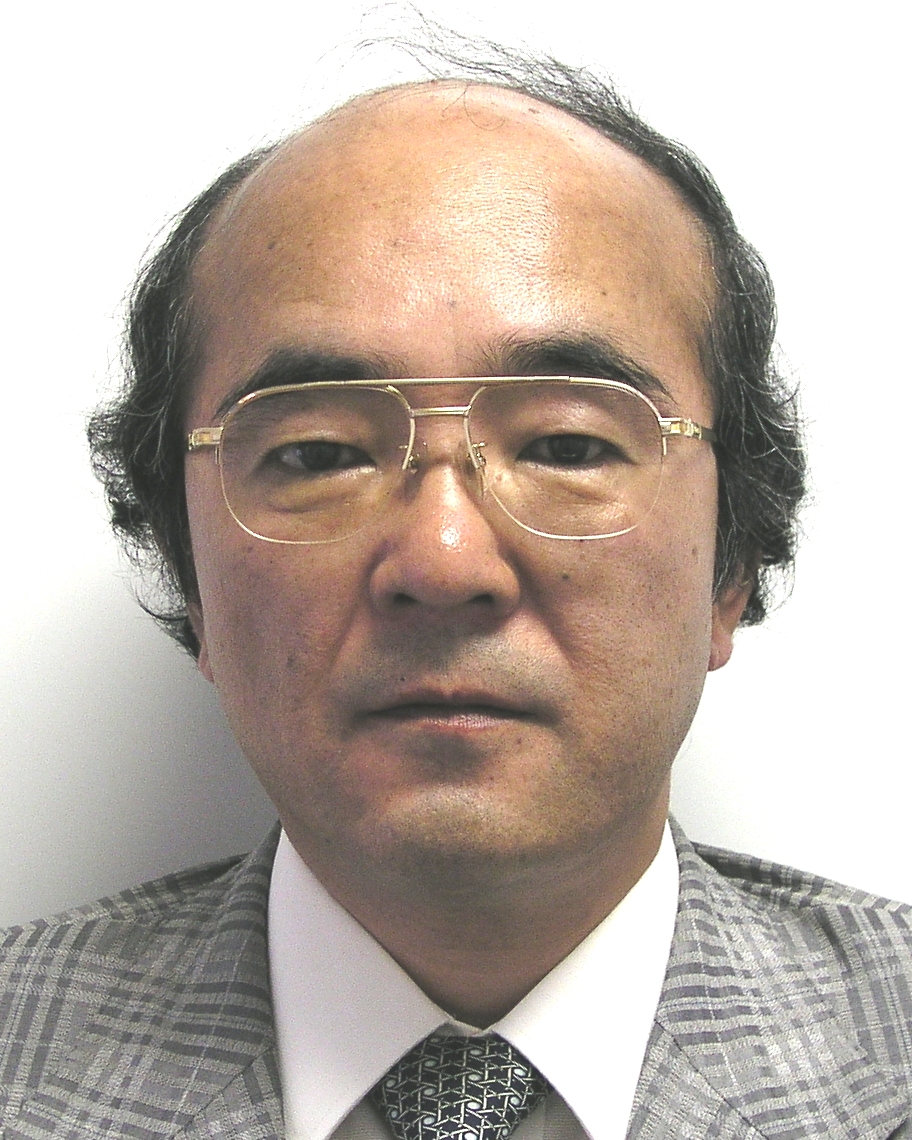}}]
{Naokazu Yokoya}
received a B.E., a M.E., and a Ph.D. in information and computer sciences from Osaka University in 1974, 1976, and 1979, respectively. He joined the Electrotechnical Laboratory (ETL) at the Ministry of International Trade and Industry in 1979. He was a visiting professor at McGill University in 1986-1987 and was a professor at Nara Institute of Science and Technology (NAIST) during 1992-2017. He is the president at NAIST since April 2017. His research interests include image processing, computer vision, and mixed and augmented reality. He is a life senior member of IEEE.
\end{IEEEbiography}

\vspace{-10.5cm}

\begin{IEEEbiography}
[{\includegraphics[width=1in,height=1.25in,clip,keepaspectratio]{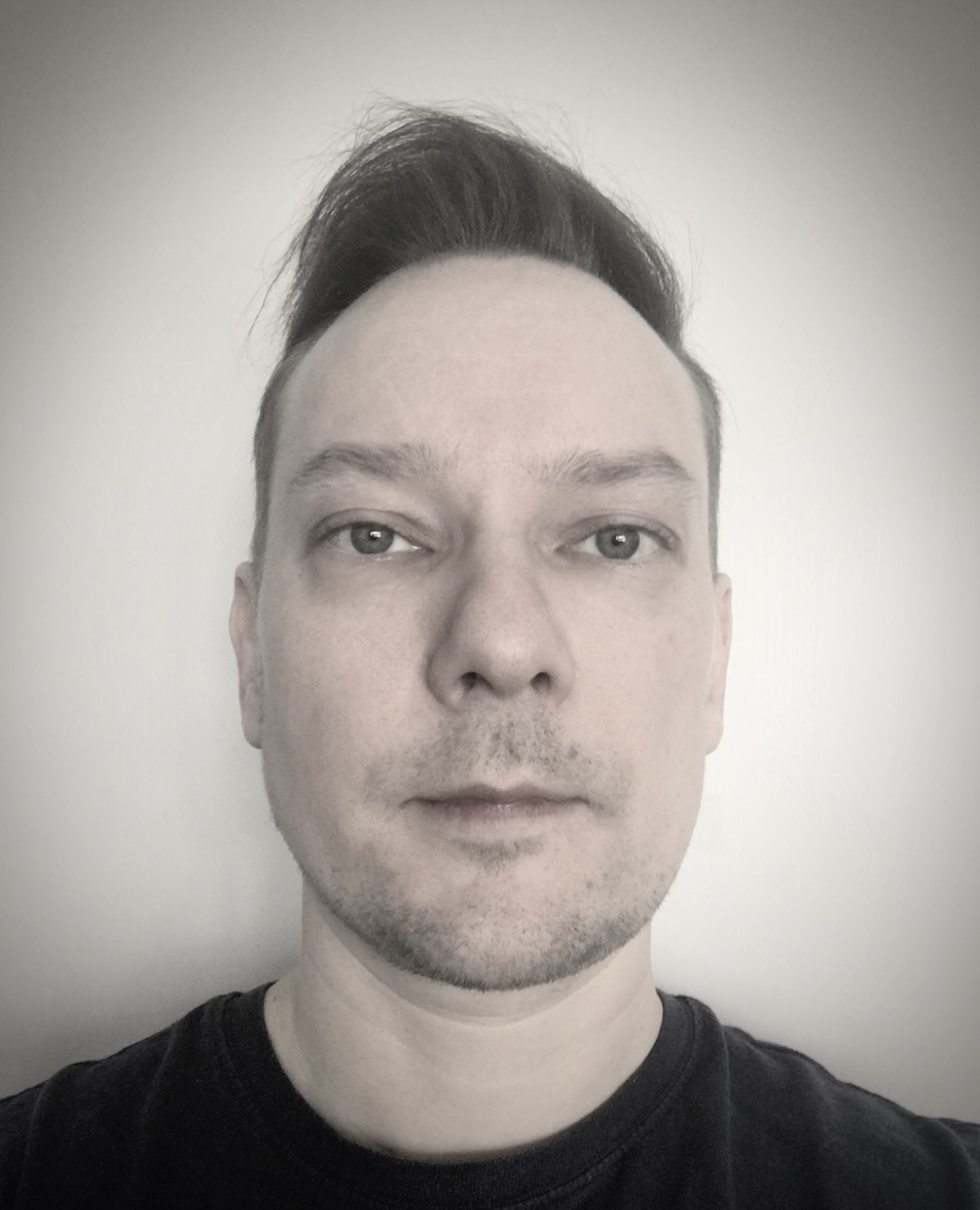}}]
{Marko Linna}
received a M.Sc. in computer science and engineering from the University of Oulu, Finland, in 2016. He is a research assistant at the Center for Machine Vision and Signal Analysis (CMVS) research center at the University of Oulu. His main research interests include computer vision.
\end{IEEEbiography}

\vspace{-10.5cm}

\begin{IEEEbiography}
[{\includegraphics[width=1in,height=1.25in,clip]{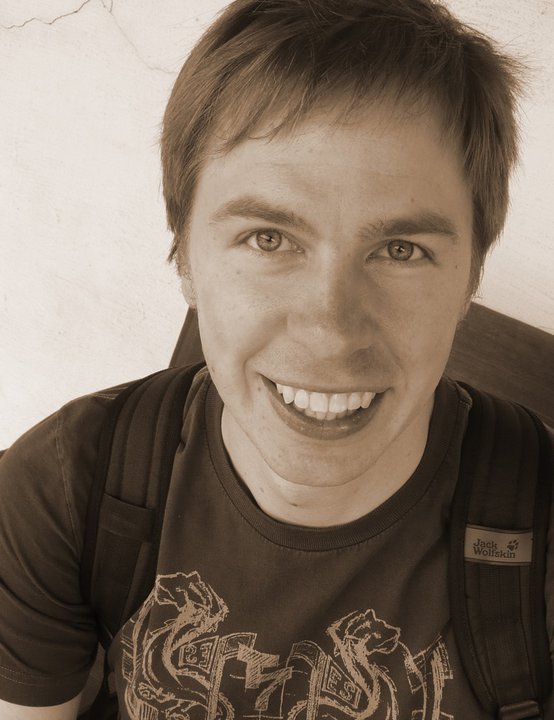}}]
{Esa Rahtu}
received a MSc in 2004 and a Doctor of Technology in electrical engineering in 2007 from the University of Oulu, Finland. His thesis was selected as the best doctoral thesis by the Pattern Recognition Society of Finland. He was also awarded a highly competitive three-year postdoctoral scholarship from the Academy of Finland. Currently, he is an assistant professor at the Tampere University of Technology in Finland. His main research interests include computer vision and machine learning with a focus on features extraction, object classification and categorization, face recognition, image retrieval, and registration and segmentation of deformable scenes. 
\end{IEEEbiography}

% if you will not have a photo at all:
%\begin{IEEEbiographynophoto}{John Doe}
%Biography text here.
%\end{IEEEbiographynophoto}

% insert where needed to balance the two columns on the last page with
% biographies
%\newpage

%\begin{IEEEbiographynophoto}{Jane Doe}
%Biography text here.
%\end{IEEEbiographynophoto}

% You can push biographies down or up by placing
% a \vfill before or after them. The appropriate
% use of \vfill depends on what kind of text is
% on the last page and whether or not the columns
% are being equalized.

%\vfill

% Can be used to pull up biographies so that the bottom of the last one
% is flush with the other column.
%\enlargethispage{-5in}

% that's all folks
\end{document}